%% file: root.tex
\documentclass[graybox]{svmult}

\input{preamble.tex}

\acrodef{DGNSS}{Differential Global Navigation Satellite System}
\acrodef{GNSS}{Global Navigation Satellite System}
\acrodef{GPS}{Global Positioning System}
\acrodef{GLONASS}{Globalnaya Navigazionnaya Sputnikovaya Sistema}
\acrodef{RTK}{Real Time Kinematic}
\acrodef{DOP}{Dilution Of Precision}
\acrodef{SNR}{Signal to Noise Ratio}
\acrodef{ENU}{East North Up}
\acrodef{IMU}{Inertial Measurement Unit}
\acrodef{NMEA}{National Marine Electronics Association}
\acrodef{CQ}{Coordinate Quality}
\acrodef{GIS}{Geographic Information System}

\usepackage{mathptmx}       %
\usepackage{helvet}         %
\usepackage{courier}        %
\usepackage{type1cm}        %
\usepackage{makeidx}         %
\usepackage{graphicx}        %
\usepackage{subcaption}
\usepackage{multicol}        %
\usepackage[bottom]{footmisc}%

\makeindex             %

\begin{document}

\authorrunning{P, Dandurand, P, Babin, V, Kubelka, P, Gigu\`{e}re and F. Pomerleau}

\title*{Predicting GNSS satellite visibility from dense point clouds}
\author{Philippe Dandurand, Philippe Babin, Vladim\'{i}r Kubelka, Philippe Gigu\`{e}re and Fran\c{c}ois Pomerleau%
	\thanks{The authors are from the Northern Robotics Laboratory, Universit\'{e} Laval, Canada.\newline
		}}
\maketitle

\abstract{To help future mobile agents plan their movement in harsh environments, a predictive model has been designed to determine what areas would be favorable for \ac{GNSS} positioning.
The model is able to predict the number of viable satellites for a \ac{GNSS} receiver, based on a 3D point cloud map and a satellite constellation.
Both occlusion and absorption effects of the environment are considered.
A rugged mobile platform was designed to collect data in order to generate the point cloud maps.
It was deployed during the Canadian winter known for large amounts of snow and extremely low temperatures.
The test environments include a highly dense boreal forest and a university campus with high buildings.
The experiment results indicate that the model performs well in both structured and unstructured environments.
}
\keywords{GNSS, GPS, lidar, RTK, DGNSS, winter, mapping, uncertainty}

\acresetall

\section{Introduction}
\label{sec:intro}

Demand for precise localization of heavy machinery is on the rise in many industries, such as agriculture, mining, and even highway truck driving.
This demand is partially driven by a strong push for \emph{Industry 4.0}, which relies on automation, real-time planning and tracking of supply chain.
It is also driven by an increased need for safety of workers. For instance, logging is the single most dangerous profession in the U.S.A.~\cite{UsBureauStats}, and could thus benefit from such an increase in vehicle automation.
Due to the large variety of outdoor environments in which heavy machinery operates (cities, open spaces, forests), one must be able to ensure that localization will be robust.
To estimate its position, an autonomous vehicle often combines measurements from different sources, such as 3D laser scanners (lidar), \ac{IMU}, cameras and \ac{GNSS} solutions.
In this paper, we focus strictly on the case where the \ac{GNSS} modality is used online, along with a 3D map generated beforehand from lidar data.

\ac{GNSS} consists of two parts: ground receivers, and constellations of satellites transmitting precise, atomic-clock-based time signal.
Each of these satellites continuously transmits a radio signal which encodes time measurements given by its on-board atomic clock.
The \ac{GNSS} receiver can estimate its distance (called a \emph{pseudo-range}) from a satellite, based on the difference between its internal clock and the received satellite clock.
With signals from at least four different satellites, it is possible to uniquely determine the position of a receiver, by solving a least-squares linear problem.
The main factors affecting the quality of a \ac{GNSS} are the number of visible satellites, and the precision of individual pseudo-range measurements.
To increase precision and robustness, modern \ac{GNSS} receivers now rely on multiple satellite constellations (e.g., \ac{GPS} from the United States, Russian \ac{GLONASS}, Galileo from the European Union and China's Beidou).
In ideal conditions, the radio signal from the satellite to the receiver would travel without interference.
In reality, this signal is affected by the Earth's atmosphere.
It also possibly reaches the receiver via multiple paths, due to reflections from objects on the Earth's surface. 
To cope with these problems, \ac{GNSS} solutions can be augmented by a correction system, becoming a \ac{DGNSS}.

\begin{SCfigure}
\centering
\includegraphics[width=0.6\linewidth]{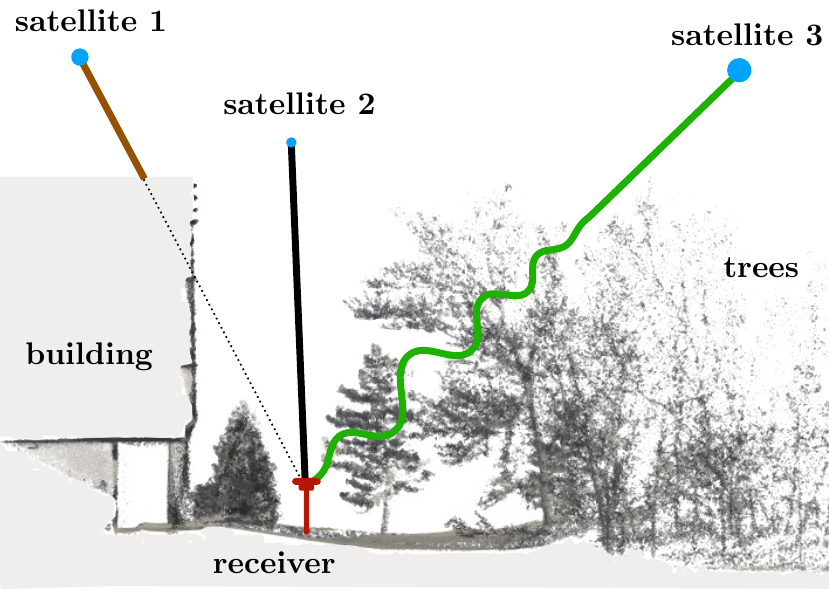}
\caption{
Side view of a complex environment with a building on the left and trees on the right. 
A \ac{GNSS} receiver is in the middle, trying to capture satellite information.
Signals from \ac{GNSS} interact differently with the environment, depending on the structure in play.
}
\label{fig:problem}
\end{SCfigure}

Although the state-of-the-art \ac{DGNSS} receivers can achieve sub-centimeter accuracy in nominal conditions, deployments in increasingly harsher environments such as forests bring new challenges.
Contrary to buildings, trees do not reflect the signals, and therefore the multi-path error (called \emph{urban-canyon effect}) is less prominent (see \autoref{fig:problem}).
Instead, trees tend to gradually absorb the signals coming from satellites \cite{PRED_LIGHT_INTER_lidar}.
Importantly, this attenuation of signals may render some satellites unusable by the receiver~\cite{GPS_forest_abso_coef}.
In such conditions, \ac{GNSS} receivers may not be as precise~\cite{GPS_perf_open_forest}, even with the correction systems~\cite{GPS_ACC_AN_FRST}.
There is therefore a strong need to be able to predict the impact of the environment on these signals, in order to assert the ability to localize.

One way to quantify the quality of a satellite signal is through its \ac{SNR}.
Research so far has focused mainly on means to either estimate the effect of the environment on the \ac{SNR} for individual satellite, or to determine the effect on the positioning as a whole.
In this paper, we propose a model that can predict the visibility of a \ac{GNSS} constellation in complex environments such as a highly dense forest with high absorption and urban canyons with high occlusion by buildings. 
To achieve this prediction, our model takes into account the geometry of both the satellite constellation and the environment. 
In particular, our approach is able to leverage local shape and density of 3D point clouds to determine which mechanism (occlusion vs. absorption) is more likely, for a given satellite.

\section{Related work}
\label{sec:relatedworks}

Several ways of estimating the impact of the environment on disrupting the satellite signal of \ac{GNSS} has been proposed in the literature.
One related line of work investigates the impact of non-absorbing obstacles such as buildings on \ac{GNSS} signals.
\citet{3D_Model_GPS_lidar} explicitly modeled the environment in 3D for the purpose of determining the availability of \ac{GNSS} signals through line-of-sight.
\ac{GIS} techniques (i.e., a combination of photogrammetry and lidar-based elevation mapping) were used to create polygonal models of buildings.
Similarly, the work of \citet{Path_Plan_UAV} shows how to use a 3D model of an urban canyon (provided by the \emph{Google Earth} application) to determine the pseudo-range error caused by the multi-path signal reception.
The same 3D model was also used to determine the distribution of the positioning error in a given area.
The knowledge of the pseudo-range error in urban canyons allows correction of the \ac{GNSS} positioning as presented in \cite{GPS_error_corr_3d}.
Their approach achieves reduction of the positioning error from \SI{9.2}{} to \SI{5.2}{\m}, using 3D models based on publicly available \ac{GIS} data.
In \cite{REFLEX_3D_NEW}, reflected satellite signals are interpreted as new satellites which add new information to the problem.
This approach was tested at a wind tunnel intake whose front wall acted as a perfect reflector due to a safety wire mesh.

These methods \cite{3D_Model_GPS_lidar,Path_Plan_UAV,GPS_error_corr_3d,REFLEX_3D_NEW} assume that a precise 3D model of the environment, typically constructed from \ac{GIS} data, is available beforehand. 
Moreover, they target exclusively urban canyons, which is too restrictive for some industries, such as forestry.
Another line of work studied the impact of \emph{signal attenuation} from absorption, with a special attention to trees.
For instance, \citet{L_band_Forest_Can} has shown that one can use the absorption coefficient index and the elevation map of a forest to estimate the effect of the vegetation on \ac{SNR} signals.
An extension of this work employed sky-oriented photos as a mean of estimating canopy closure in pine forests~\citep{ESTM_PINE_PHOTO}. 
This method only estimates the signal loss, but cannot determine how many satellites would be viable. Therefore, it is not well suited for satellite visibility prediction. 
Similarly, \citet{Forest_Canopy_GPS_perf} established a relationship between the fragmentation of sky pictures and the accuracy of the positioning.
However, this method focused on tree canopy using stationary analysis without modeling the effect of structures such has buildings. It also only used photos to determine obstacles. %
In our work, we extend this approach by explicitly modeling the effect of structures that exhibit a masking effect, from their uniform flat surfaces.

While some methods estimate the pseudo-range error from an a priori 3D model of the environment, others exploit 3D point clouds to determine whether or not a given satellite signal is reachable by the \ac{GNSS} receiver~\cite{GPS_remove_3d_points}.
The line-of-sight between the satellites and the receiver is then simply determined by using a ray tracing algorithm.
By identifying these satellites, the authors were capable of estimating offline a more realistic confidence on the positioning.
This method, however, exhibits problems with large trees since they do not necessarily block the signals entirely, but rather may absorb some of it.
Therefore, simply removing a satellite because of tree occlusions makes their model under estimate the positioning confidence.
\citet{Point_cloud_GNSS_Sky_view} proposed to use a 3D point cloud to generate a binary mask of the sky to remove occluded satellites.
A similar method \cite{GPS_GLONASS_INFRA} has been used in an urban environment, using infrared images taken at night, to estimate satellite masking from buildings.
Again, if a satellite lays behind \emph{any} obstacle (building or tree), it is entirely masked.
This binary approach is not well suited for the highly dense forest environment. 
Indeed, most of the satellites would be deemed as masked, given the fact that the majority of the sky is occluded by the canopy.
Our approach is similar to the last two, but also take into consideration the type of structures affecting the occluded satellites.
More precisely, we take into account that trees do not block the signal completely, but rather decrease their \ac{SNR}.

\section{Theory}
\label{sec:theory}
Our goal is to predict how the environment influences the ability of a \ac{GNSS} receiver to localize itself. More precisely, we look at modeling the environment's interaction with the signal of each satellite, from the ground receiver perspective. 
To this effect, we leverage \ac{GNSS} constellation configuration information and a representation of the environment, namely a large 3D point cloud map of an area of interest, to predict an effective number of satellites available for localization.

\subsection{Processing \ac{GNSS} information}

Ground receivers produce information, in the form of \emph{sentences}, under the format of \ac{NMEA}. We use three specific sentences to understand the effect of the environment on the receiver:
\begin{enumerate*}
    \item \texttt{RMC} for the estimated position of the receiver,
    \item \texttt{GST} for statistics on the positioning of the receiver, and
    \item \texttt{GSV} for information on individual satellite. 
\end{enumerate*}
From the \texttt{GST} sentence, one can extract a positioning metric called \ac{DOP}. 
This metric is a unitless indicator of the confidence the receiver has in a positioning output. 
For example, a \ac{DOP} of 1 is considered as ideal, but one higher than 20 signifies poor positioning accuracy.
With the \texttt{GSV} sentence, one obtains a satellite's position in the sky (elevation and longitude) as well as its \ac{SNR}. 
A good \ac{SNR} would be around 45 dB, while a bad one would be around 30 dB.
This information allows an understanding of what is happening to the signals, particularly when measuring how different environment are impacting it.
To make sure we get a good positioning, the mobile receiver is corrected by a reference station.
The correction comes from using \ac{RTK}, a reference station measures errors, and that station, then transmits corrections to the mobile receiver.
We get the satellite constellation from the reference station used for the \ac{RTK} correction.

\begin{figure}[htbp]
\centering
\includegraphics[width=0.9\textwidth]{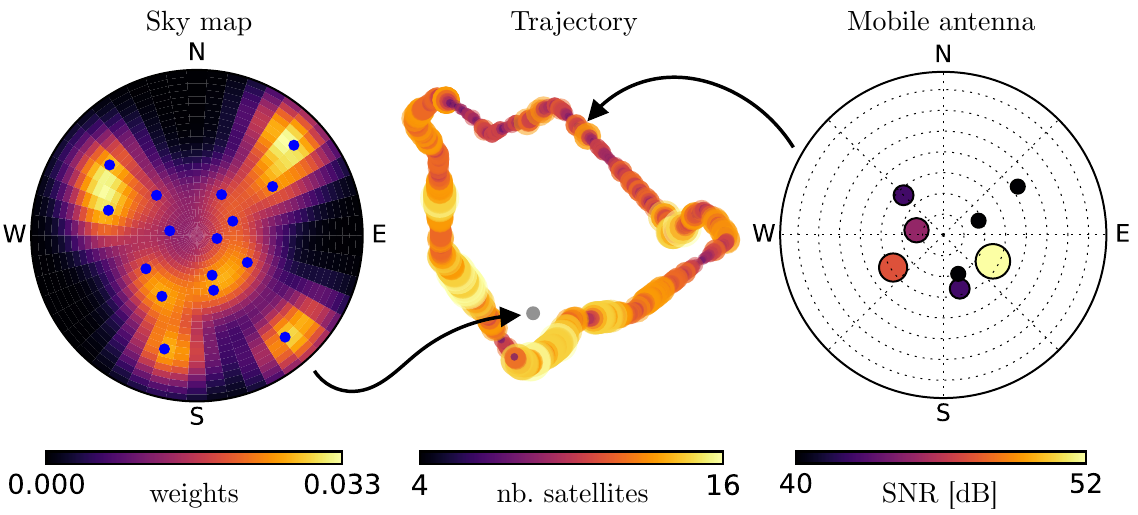}

\caption{
Different uncertainty information accessible through \ac{NMEA} sentences. 
\emph{Left}: individual satellite positions from a reference antenna installed in an open area and used as reference to build the sky map $\bm{S}$.
\emph{Middle}: trajectory of a moving receiver passing from open sky to a forest, size and color coded with the number of visible satellites at each position.
The large variation of the latter is the justification for our approach.
\emph{Right}: individual satellite position with larger dot representing higher \ac{SNR} for a single position of the moving receiver.
}
\label{fig:satelliteInfo}
\end{figure}

In our model, we represent a satellite constellation as a Gaussian mixture model, with one Gaussian per satellite. 
Its mean represents the position of a satellite, while its variance $\sigma$ captures the effect of the \ac{GNSS} act as a radio wave source.
We further process this representation as a discretized hemispherical projection of the sky $\bm{S}$, with each cell $s_{ij}$ having a resolution of $e \times l$ (i.e., division on longitude and elevation). 
Given that Gaussians are used, this \emph{sky maps} has the property that summing all cells $s_{ij} \in \bm{S}$ gives the total number of satellites $v$ that are visible at this time and place.
An example of such hemispherical projection $\bm{S}$, 
and the variation of visible satellites along a trajectory can be seen in the \autoref{fig:satelliteInfo}.

\subsection{Processing 3D Points}
\label{sec:Processing3DPoints}
The second source of information comes from a dense 3D representation of the environment in the form of a map $\mathbf{M}$, generated offline.
One issue with mobile Lidars is that they produce an uneven distribution of points when registered together to form such a 3D map.
To mitigate this, we used an octree to produce a uniform distribution of points by keeping a single point per bounding box of $d_{\text{box}}$ wide.
Since the environment can either absorb (e.g., forest) or block (e.g., buildings) signals
our model has to treat these sources of obstruction differently.
We rely on the shape of the distribution of neighbouring points to determine if a given obstacle is most susceptible to cause absorption or occlusion.
More precisely, for each point, we compute the covariance of its $k_\text{nn}$ nearest neighbours.
If the distance between a tested point and the mean $\mu_\text{nn}$ of its surrounding points is greater than $d_{\text{nn}}$, this point is on a corner or at the periphery of the map. 
We consider that for these points, the covariance computation is invalid, and they are thus removed.
For the remaining valid covariances, we then sort their eigenvalues $\lambda_i$, with $\lambda_1 < \lambda_2 < \lambda_3$.
By analyzing their relative ratios, we defined local features that try to capture the level of `unstructureness' $u \in [0,1]$, `structureness' $s \in [0,1]$, and the spherical level $\delta \in [-1, 1]$ of each point. These features are defined as: 
\begin{equation}
\label{eq:structureLevel}
u = \frac{\lambda_1}{\lambda_{3}}
\text{, \hspace{0.5cm} }
s = \left( \frac{\lambda_{2}}{\lambda_{3}} \vphantom{\frac{\lambda_2 - \lambda_1}{\sqrt{(\lambda_{2}^{2} + \lambda_{1}^{2})}}} \right)
 \left( \frac{\lambda_2 - \lambda_1}{\sqrt{(\lambda_{2}^{2} + \lambda_{1}^{2})}} \right)
 \text{, \hspace{0.5cm} and \hspace{0.5cm}}
 \delta = u - s .
\end{equation}
We can observe from \autoref{eq:structureLevel} that, as $\lambda_1$ approaches $\lambda_3$, the value of $u$ will be close to one. This will be the case, for spherical distributions, typical of objects present in unstructured environments. 
A geometrical interpretation is less straightforward for $s$.
When looking at the first ratio for $s$, as $\lambda_2$ is approaching $\lambda_3$, the covariance tend to assume a circular shape, while the second ratio will tend to one if $\lambda_1$ is a lot smaller than $\lambda_2$ (i.e., a flat shape).
If the local geometry is close to a line, then both $u$ and $s$ will be near zero.
By subtracting these two values, we can represent the spherical level on one axis $\delta$ with values close to minus one being plans, around zero being edges, and close to one being a diffuse sphere.
\autoref{fig:pointCloudInfo} (\emph{left}) shows an example of the spherical level $\delta$ in an environment with buildings and trees.

\subsection{Predicting the number of satellites}

\begin{figure}[htbp]
\centering
\begin{minipage}[t]{0.9\textwidth}
\begin{tabu}{X[c] X[c] X[c] X[c]}
Point cloud & Binary mask & Spherical level $\delta_\text{med}$ & Signal reduction $p(\cdot)$
\end{tabu}
\includegraphics[width=0.24\linewidth, trim={11mm 5mm 12mm 0mm}, clip]{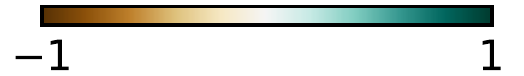}
\includegraphics[width=0.24\linewidth, trim={11mm 5mm 12mm 0mm}, clip]{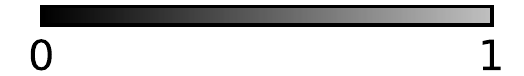}
\includegraphics[width=0.24\linewidth, trim={11mm 5mm 12mm 0mm}, clip]{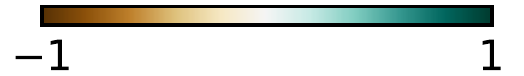}
\includegraphics[width=0.24\linewidth, trim={11mm 5mm 12mm 0mm}, clip]{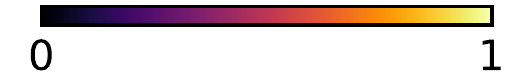}
\includegraphics[width=0.24\linewidth, trim={5mm 5mm 5mm 4mm}, clip]{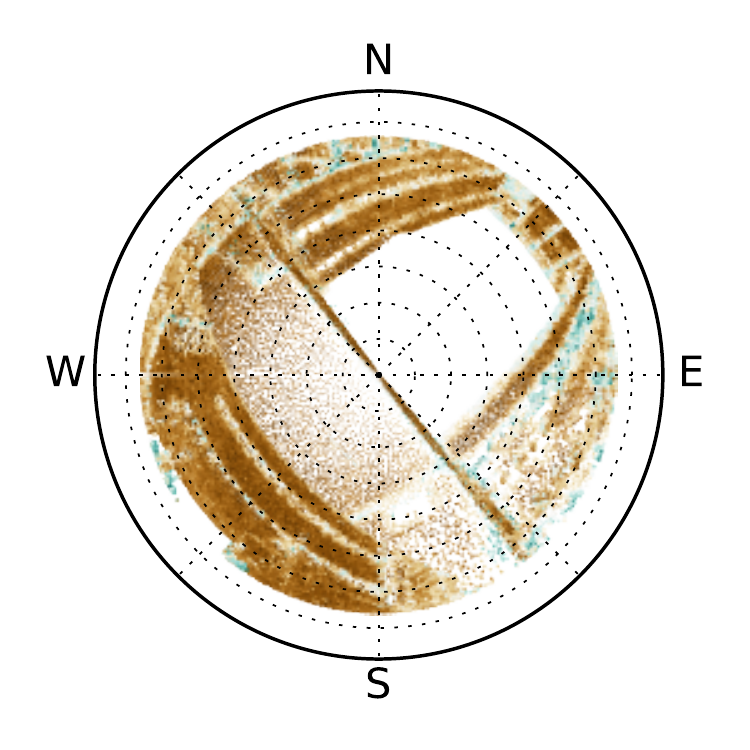}
\includegraphics[width=0.24\linewidth, trim={5mm 5mm 5mm 4mm}, clip]{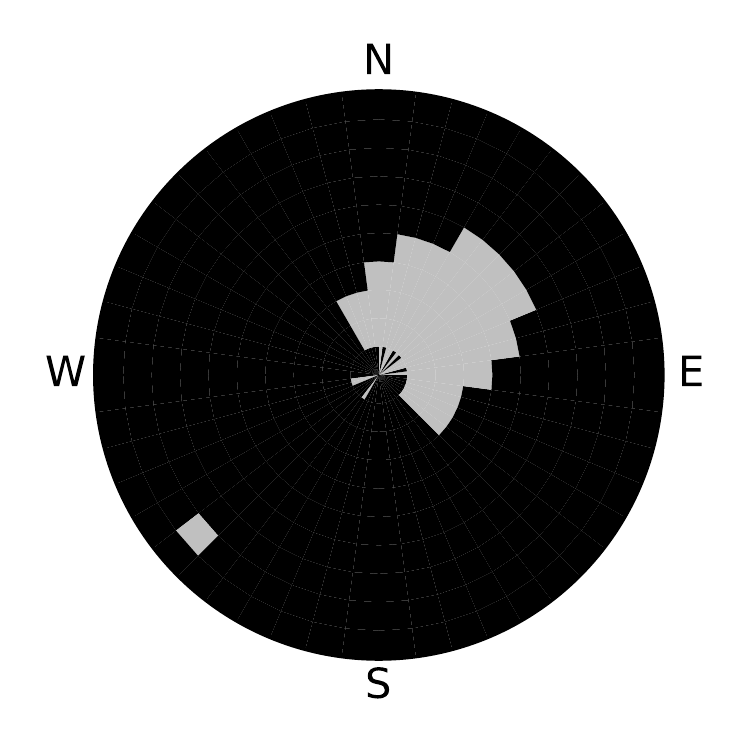}
\includegraphics[width=0.24\linewidth, trim={5mm 5mm 5mm 4mm}, clip]{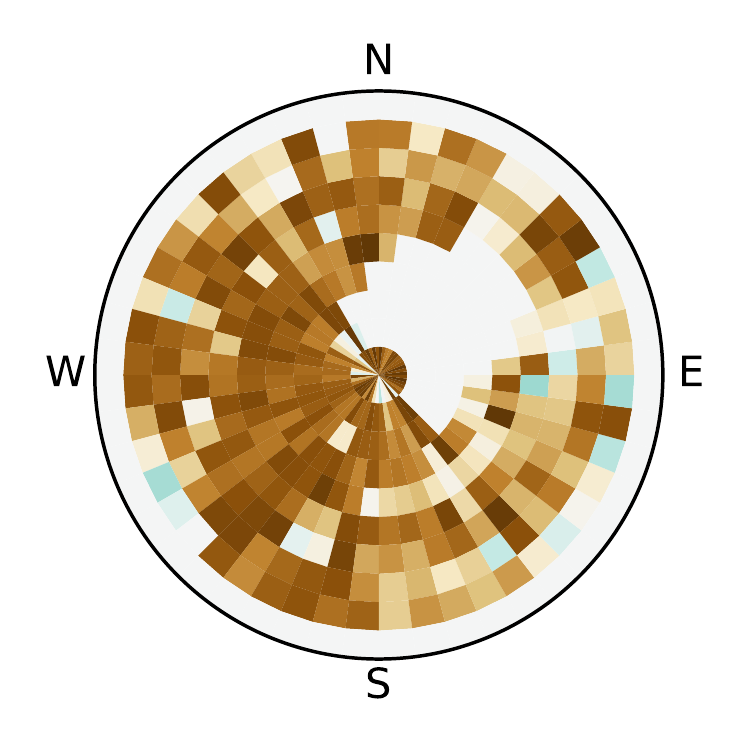}
\includegraphics[width=0.24\linewidth, trim={5mm 5mm 5mm 4mm}, clip]{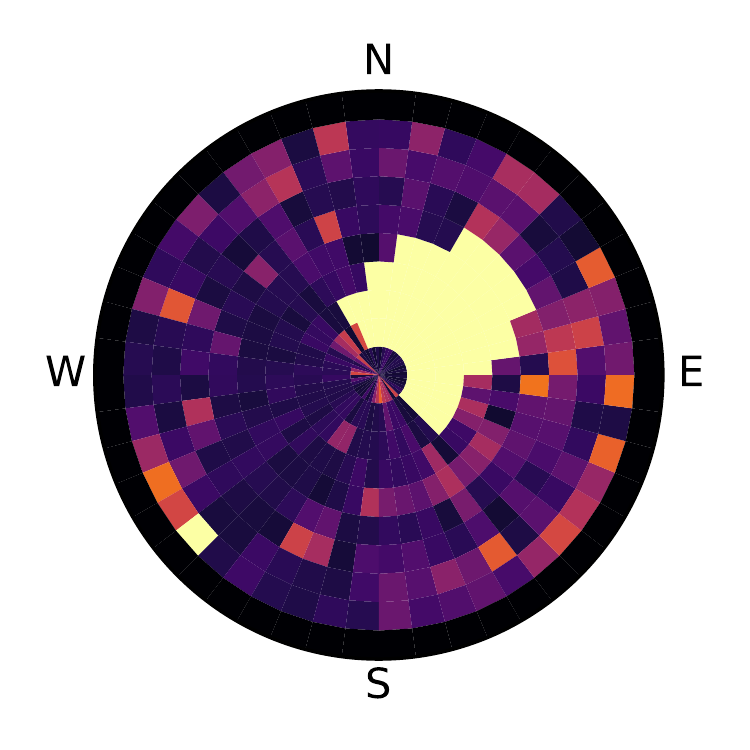}
\includegraphics[width=0.24\linewidth, trim={5mm 5mm 5mm 4mm}, clip]{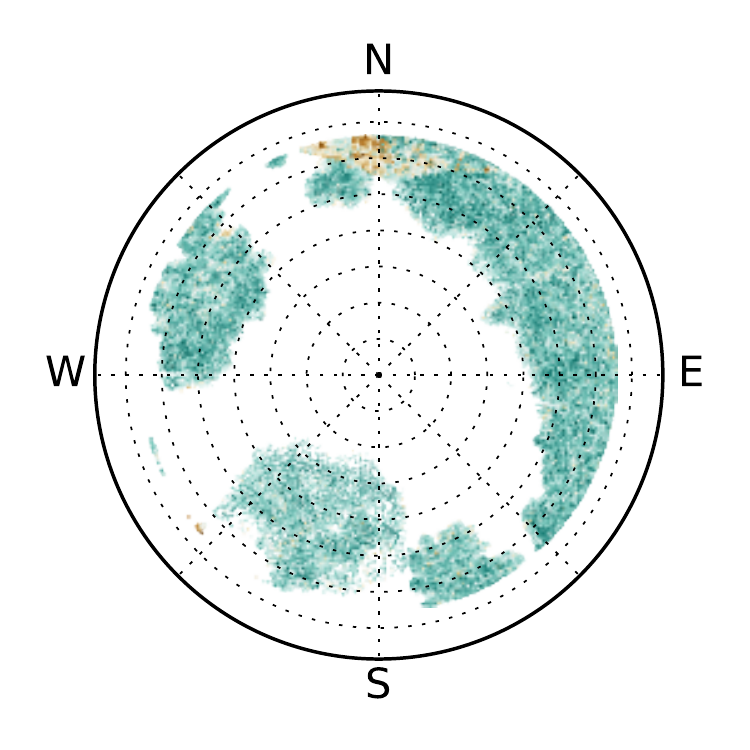}
\includegraphics[width=0.24\linewidth, trim={5mm 5mm 5mm 4mm}, clip]{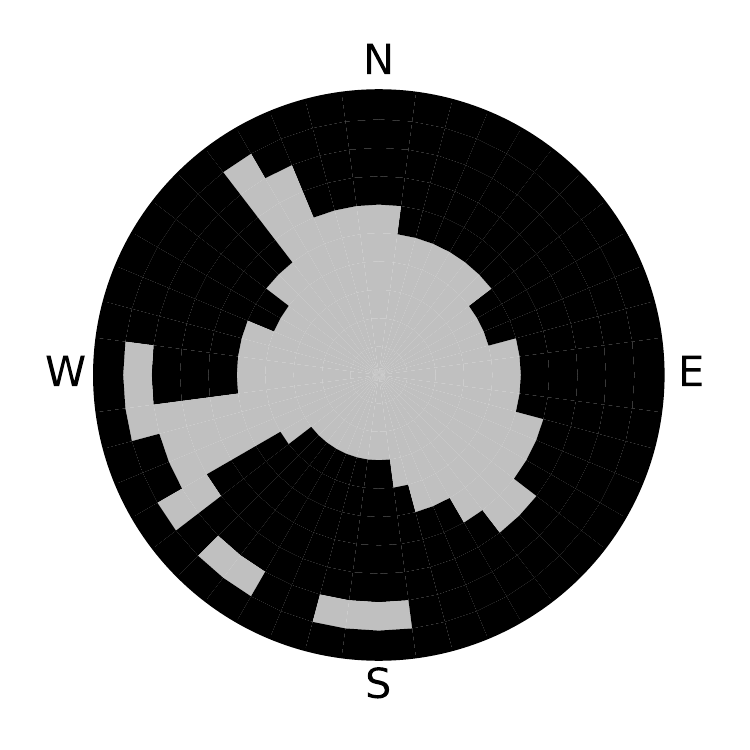}
\includegraphics[width=0.24\linewidth, trim={5mm 5mm 5mm 4mm}, clip]{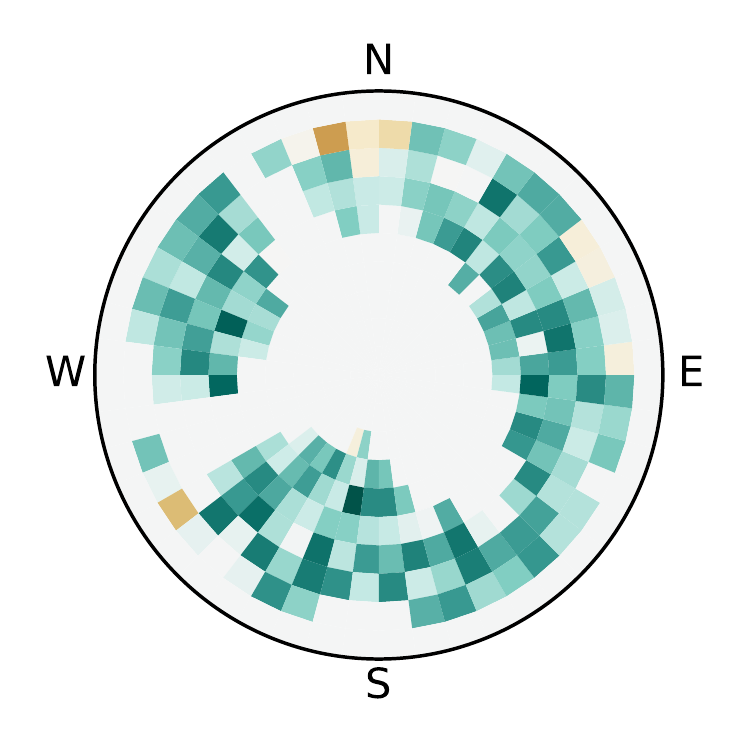}
\includegraphics[width=0.24\linewidth, trim={5mm 5mm 5mm 4mm}, clip]{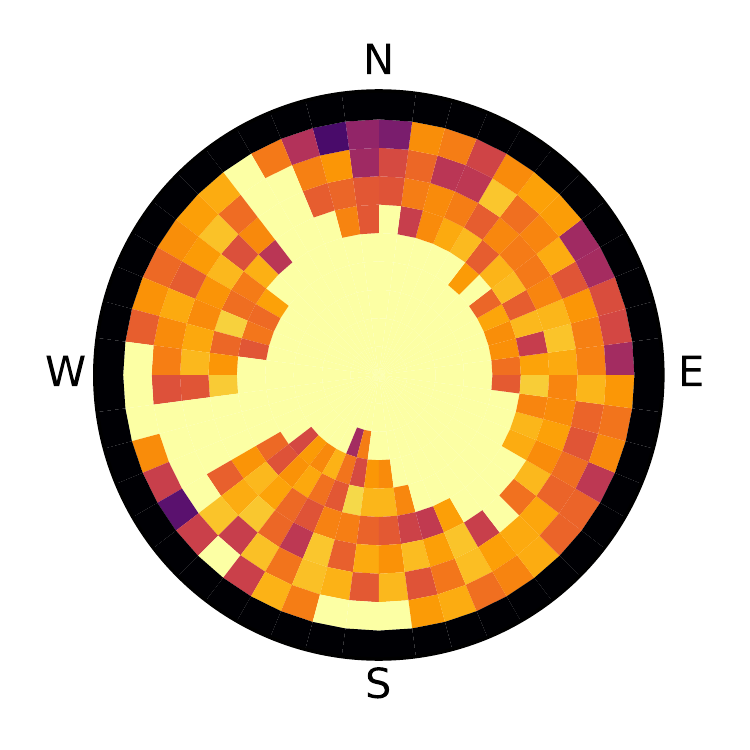}
\caption{Example of signal reduction. 
\emph{Top row}: a structured environment with a building covering  part of the sky.
\emph{Bottom row}: an unstructured environment from a path in a forest.
In the point cloud, the points have been colored with their spherical level value $\delta$ to show the difference between structured and unstructured environment. Those values are reflected in the spherical level histogram and have an effect on the signal reduction.
}
\label{fig:histograms}
\end{minipage}
\end{figure}

For a given point on the ground and its normal vector $\bm{n}$, we compute the estimated number $\hat{v}$ of visible satellites.
To do so, we developed a hemispherical and discretized reduction model $\bm{P}$ of the sky perceived by a receiver at this pose, aligned with the normal $\bm{n}$.
It is built with the same resolution as our satellite map $\bm{S}$.
For each cell $p_{ij} \in \bm{P}$, two statistics are extracted from the 3D points of the map $\bm{M}$ falling into its angular bounding box.
The first one, $\delta_\text{med}$, is the median value of the $\delta$'s computed with \autoref{eq:structureLevel}, for all the points within this cell. This variable $\delta_\text{med}$ tries to capture the type of interaction (occluding vs. absorbing) a satellite signal would experience, when passing through this cell and towards the receiver.
The second statistic is the number $m$ of points in this cell. This number is independent of the lidar scanning density, due to the voxel filtering described in \autoref{sec:Processing3DPoints}. 
We propose to model the signal reduction $p(\cdot) \in [0,1]$ as a function that transition smoothly from occlusion to absorption:
\begin{equation}
\label{eq:modelTransition}
p(\delta_\text{med}, m) = \underbrace{ \Big(1 + \exp{\big(-\alpha(\delta_\text{med} - \beta) \big)} \Big)^{-1} }_{\text{from occlusion to absorption}}
\underbrace{ \exp (-\gamma m) \vphantom{\Big(^{-1}}}_{\text{absorption model}}
.
\end{equation}
The parameter $\alpha$ controls the slope of the transition, $\beta \in [-1, 1]$ the position of the transition, and $\gamma$ is the absorbance. 
For cells containing mostly occluding obstacles (such as in structured environments), $\delta_\text{med}$ is generally around -1 and the first term (sigmoid function) will be near 0. This captures the fact that when going through highly structured obstacles, the \ac{GNSS} signal is fully absorbed, i.e. $p\to0$.
For diffused obstacles, $\delta_\text{med}\approx1$, and the first term will be close to 1. 
Consequently, the signal is reduced following an exponential decay function, based on the number of points in a cell $m$ (i.e. $p$ would be approximately equal to $\exp (-\gamma m)$, the absorption model).
To let the signal go through when there are very few points, a binary mask $\bm{B}$ is applied.
Each cell $b_{ij} \in \bm{B}$ equals 1 if the number of points $m$ is lower than $m_\text{occ}$, and 0 otherwise.
By comparison, the solution proposed by \citet{GPS_remove_3d_points} is equivalent to masking a cell if the number of points $m$ is larger than 0.
\autoref{fig:histograms} gives a visual intuition of how $p(\cdot)$ is computed for two different types of environment.
Finally, we multiply our satellite representation $\bm{S}$ with the maximum of our reduction model $\bm{P}$ and our binary mask $\bm{B}$ and sum over all cells to predict the number of satellites viewed $\hat{v}$ following
\begin{equation}
\label{eq:prediction}
\hat{v} = \sum_i \sum_j s_{ij}\max (p_{ij}, b_{ij})\,
\text{, \hspace{0.5cm} with \hspace{0.5cm} } \hat{v} \leq v.
\end{equation}

\subsection{Extrapolation to Any Position on the Ground}

\begin{figure}[htbp]
\centering
\includegraphics[width=0.9\linewidth]{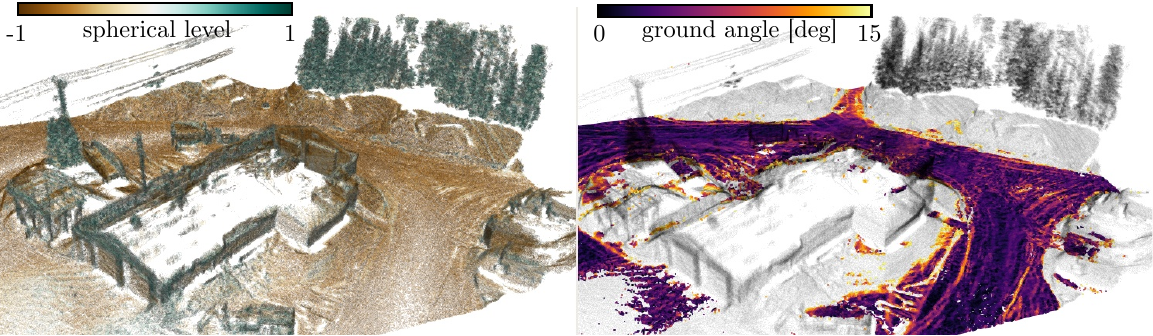}
\caption{
Different information extracted from the geometry of a point cloud for the absorption model and the ground segmentation. 
\emph{Left}: side view of a map with color representing the level of structure.
\emph{Right}: from the points strongly defined as structured, a subset of points with surface normal vector pointing upward are labeled as the ground.
}
\label{fig:pointCloudInfo}
\end{figure}

As a last processing step, we want to predict the number of satellites for all potential places in a map $\bm{M}$ where a virtual receiver could stand.
\autoref{fig:pointCloudInfo} (\emph{right}) shows an example of the segmentation of the potential positions (i.e. the ground).
For each point, the spherical level $\delta$ and its eigenvector associated to $\lambda_1$ (i.e., the surface normal vector $\bm{n}$) is used to segment the ground following the constrains $\delta < \delta_\text{ground}$ and $\arccos(n_z) < \epsilon $. 
Because the ground is not always flat, we use the surface normal $\bm{n}$ to position the virtual receiver at the proper angle and height.
This step is important to properly control the field of view of the virtual receiver in situations where a robot is heavily tilted and the antenna would point toward obstacles. 

\section{Experimental Setup and Methodology}
\label{sec:exp_setup}

\begin{figure}[htbp]
\centering
    \includegraphics[height=80px]{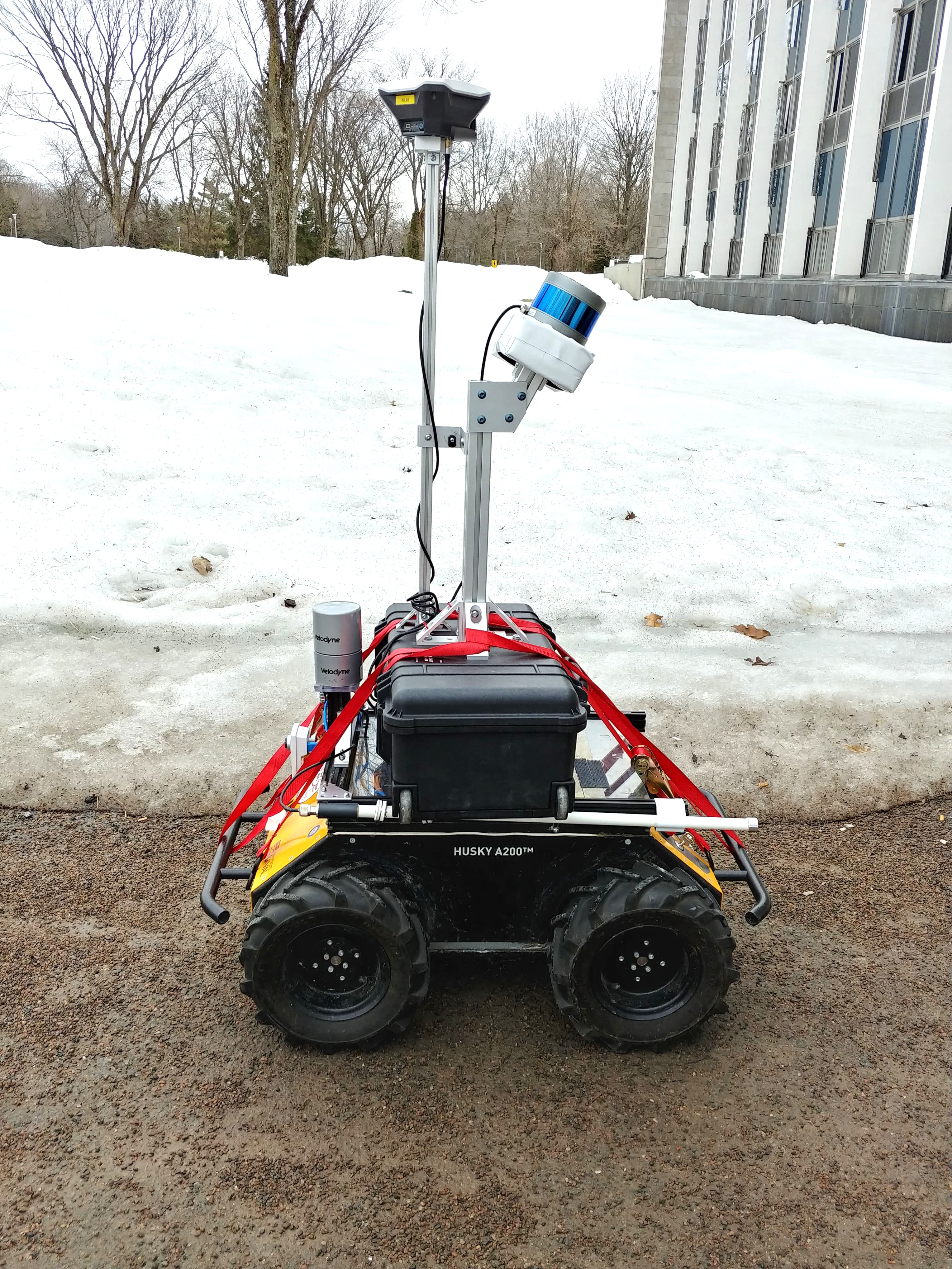}
    \includegraphics[height=80px]{figs/fig5/skidoo.pdf}
    \includegraphics[height=80px]{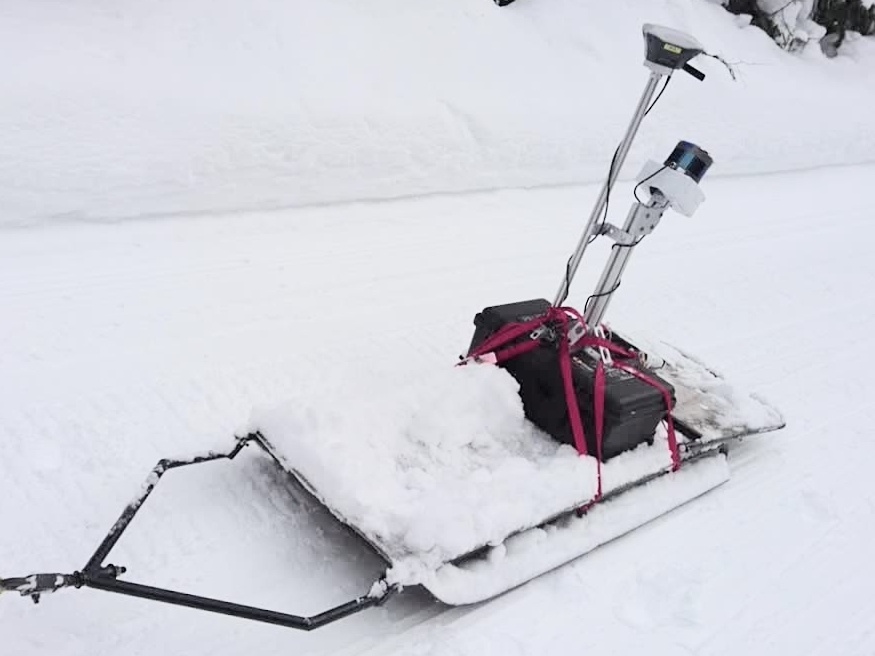} 
    \caption{
    \emph{Left}: Rack mounted on a small robot for easy terrain (e.g. parking lot, sidewalks).
    \emph{Middle}: Rack pulled by a snowmobile for maps on harden snow. 
    \emph{Right}: In last resort, the rack could be pulled by humans, which has proven to be useful on narrow/dense paths in the forest.
    }
    \label{fig:pelicanMobile}
\end{figure}

To test and validate our prediction model, we designed a mobile data collection platform composed of an \ac{IMU}, a 3D lidar, along with a mobile and a fixed \ac{GNSS} antennas.
The \ac{IMU}, an \emph{Xsens Mti-30}, was used for the map $\bm{M}$ generation and the state estimation.
The lidar, a \emph{Robosense RS-LiDAR-16}, was mounted with a tilt angle of \SI{27}{\degree} to ensure that we can capture vertical structures as high as possible. 
This was necessary to properly generate a view of the environment above the antenna via the map $\bm{M}$.
Our algorithm is independent of the mapping and localization sensors used, as long as accurate and dense 3D maps can be produced from them.
Both \ac{GNSS} antennas are model \emph{Emlid Reach~RS+}, and coupled together to provide an \ac{RTK} solution.
We used the default factory settings for which the receiver units ignored satellites beneath an elevation of \SI{15}{\degree} above the horizon or having an \ac{SNR} lower than \SI{35}{\dB}.
The mobile receiver was placed one meter above the platform to prevent obstructions from the platform itself.
As for the static receiver, it was mounted on a tripod and placed at a location having an open view of the sky.
All the mobile sensors were connected to a computer inside a \emph{Pelican} case.
This waterproof case protected the electronics, 
while serving as a mechanical mount to hold the sensors.
This rugged and compact platform, thus allowed for field deployments in harsh winter conditions. 
It also allowed for various means of transport, as displayed in \autoref{fig:pelicanMobile}, depending on terrain complexity.

\begin{SCfigure}%
\centering
        \includegraphics[height=90px]{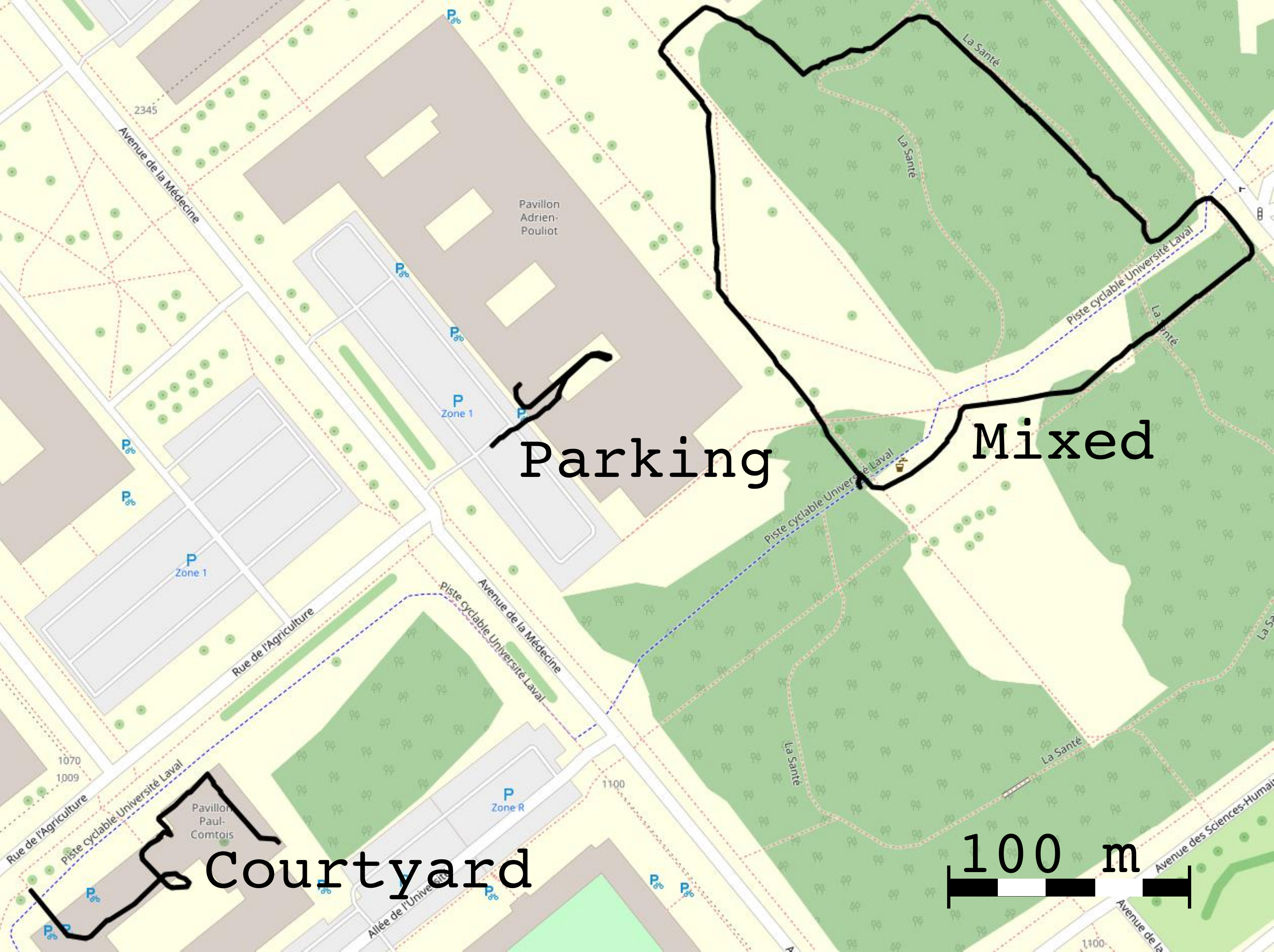}
        \includegraphics[height=90px]{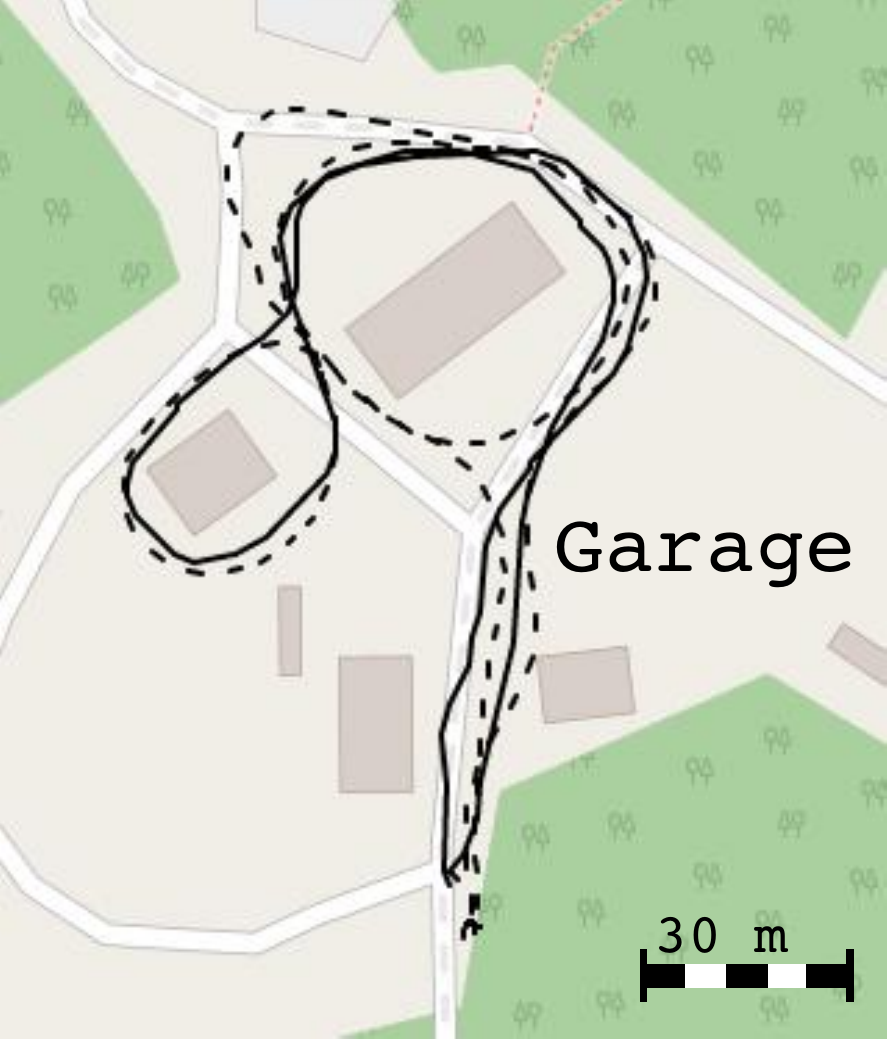}
    \caption{ Overview of all trajectories used during our experiments, all of them in Quebec, Canada.
    \emph{Left}: three trajectories on the campus of Laval University.
    \emph{Right}: two trajectories in the Montmorency Forest. The dashed line being the validation trajectory \texttt{Garage-2}.
    \textcopyright OpenStreetMap contributors.
    }
\label{fig:trajectories}
\end{SCfigure}

Five data gathering campaigns, as shown in \autoref{fig:trajectories}, were conducted on the campus Laval University and its remote research forest, the Montmorency Forest.
All environments involved in the experiments were covered by a thick layer of snow and targeted variable numbers of satellites, from zero to 20.
Some of the trajectories, namely \texttt{Parking} and \texttt{Courtyard}, are within well-structured environments.
Both of these trajectories passed under covered porches to reach interior courtyards, creating situations where few satellites are visible.
Another trajectory, \texttt{Mixed}, explored an unstructured environment. It started in an open field with sparse large trees on each side of a path and eventually traversed a deciduous forest, which is sparse during the winter with all the leaves being felt.
Finally, the last trajectory \texttt{Garage-1} was in the Montmorency Forest circling around small storage buildings. The surrounding area was a coniferous forest, which stays very dense during winter.
We adjusted our parameters based on \texttt{Garage-1}, \texttt{Mixed}, \texttt{Parking} and  \texttt{Courtyard}. 
We used \texttt{Garage-2} as our validation dataset by going back 4 weeks later for another trajectory in the same area so the satellite constellations would be different.
Finally, to support a better reproducibility of our results, we list all of the parameters used during our experiments in \autoref{tab:parameters}.
All maps and trajectories depictions follow the \ac{ENU} geographical coordinate system.

\begin{table}
\caption{List of parameters used for our results.}
\begin{tabu}{X[3] X[c] c }
\toprule
\emph{Description} & \emph{Parameter} & \emph{Experimental value} \\
\midrule
Variance on satellite angular positions & $\sigma$ & \SI{12.5}{\degree} \\
Octree bounding box size  & $d_{\text{box}}$ & \SI{0.1}{m} \\
Number of nearest neighbours & $k_\text{nn}$ & 50 \\
Maximum offset from the mean nearest neighbours & $d_\text{nn}$ & \SI{0.25}{m} \\
Maximum structured value for the ground & $\delta_\text{ground}$ & -0.6 \\
Maximum angle for ground classification & $\epsilon$ & \SI{10}{\degree}\\
Resolution of the angular histogram  & $e \times l$ & \SI{7.5}{\degree} $\times$ \SI{9}{\degree} \\
Minimum number of points for occupancy & $m_\text{occ}$ & 5 \\
reduction model & $\alpha, \beta, \gamma$  & 4,0.25,10\textsuperscript{-10}\\
\bottomrule
\end{tabu}
\label{tab:parameters}
\end{table}

\section{Results}
\label{sec:results}

First, we evaluated the capacity for our reduction model $p(\cdot)$ to adapt from structured to unstructured environments, by looking at specific trajectories and comparing with the current literature.
Then, we investigated the stability of our model through four datasets.
Finally, we validated our extrapolation of the number of satellites on all points from the ground with a separate dataset.

\subsection{Adaptability to the Environment Geometry}
\begin{figure}[htbp]
    \centering
    \includegraphics[width=0.45\textwidth, trim={15mm 15mm 15mm 15mm}, clip]{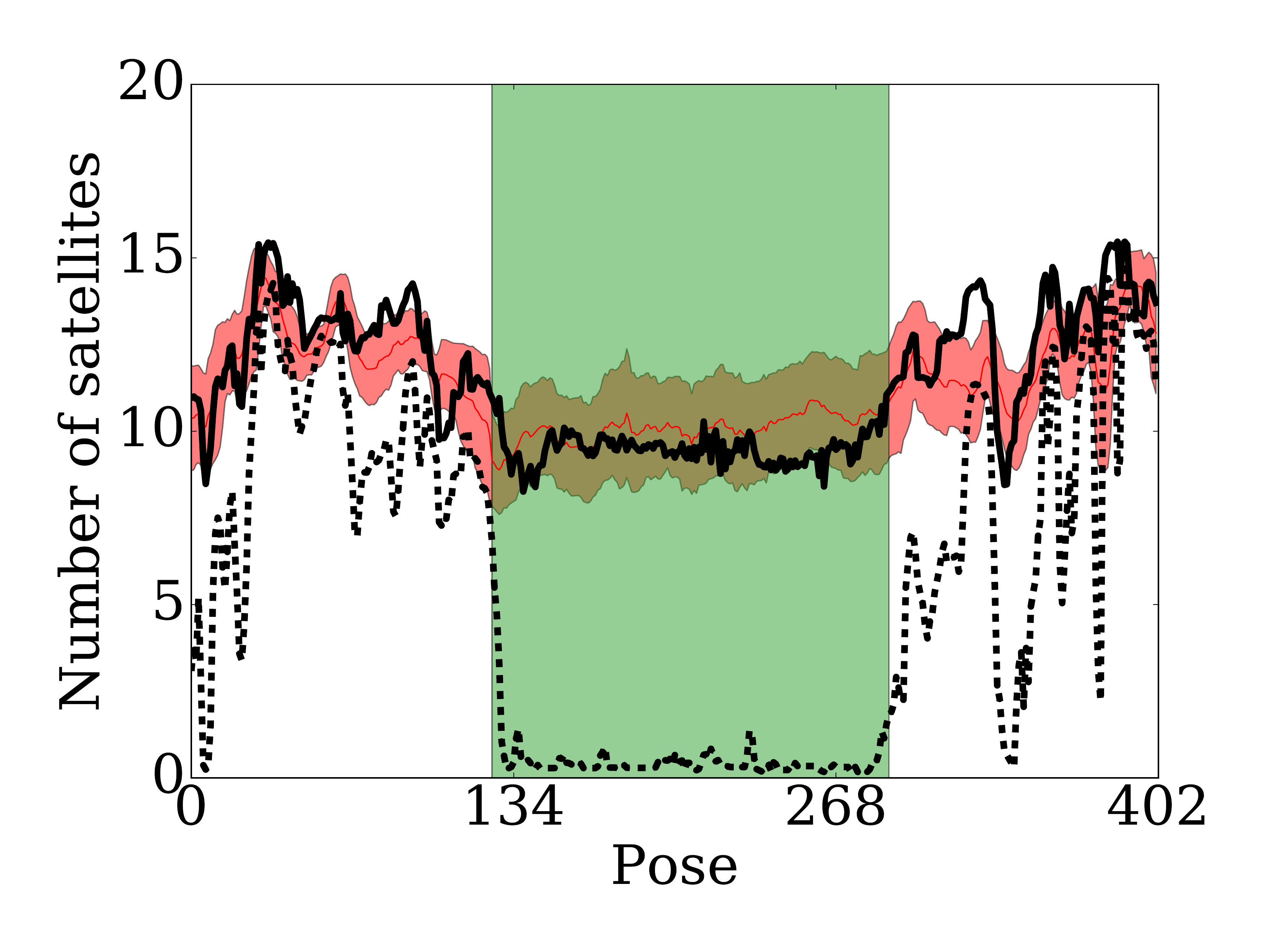}
    \includegraphics[width=0.45\textwidth, trim={15mm 15mm 15mm 15mm}, clip]{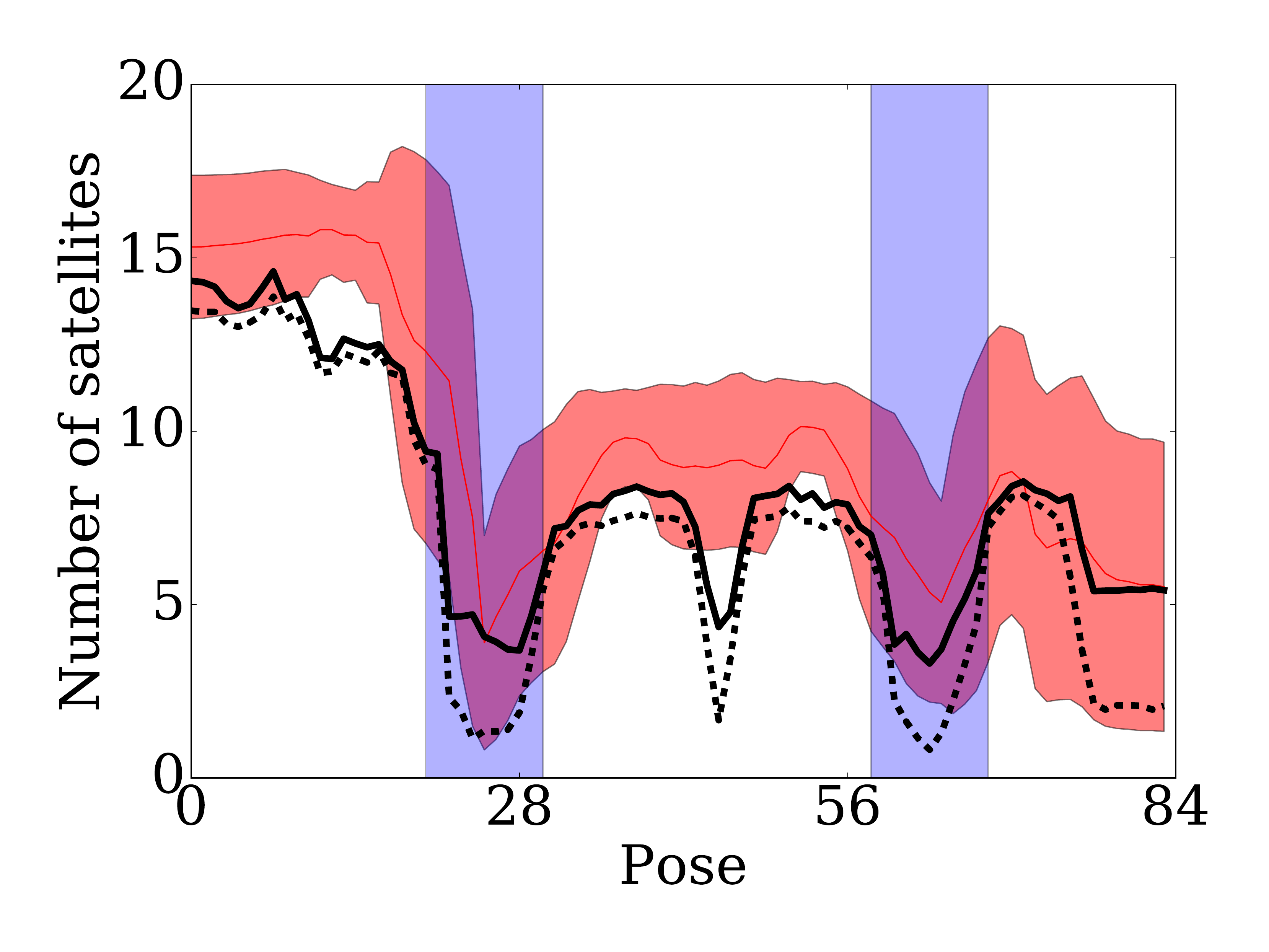}
    \caption{Number of satellites actually perceived (red line) compared to our model (black line) and the model of  \citet{GPS_remove_3d_points} (dashed black line).
    The red shaded area corresponds the standard deviation from an averaging window of \SI{5}{\m}
    \emph{Left}: trajectory from \texttt{Mixed} passing through highly dense forest that covers the sky.
    \emph{Right}: trajectory from \texttt{Parking} passing through structured environment that blocks signals.
    }
    \label{fig:traj_perf}
\end{figure}

As a baseline, we used a binary removal of the satellites as in \citet{GPS_remove_3d_points}.
This baseline was chosen because it also employs point clouds for satellite occlusion prediction.
By comparing the \texttt{Mixed}, \texttt{Parking} and the \texttt{Garage-1} for the Montmorency Forest, we determined that our model performed best in dense environments.
From \autoref{fig:traj_perf} (left), we can clearly see that for areas covered by trees (the green region), our model can predict the number of visible satellites, while the forest caused too much obstruction for the binary removal to be effective.
This is why an absorption model is appropriate for highly dense forest environment.
The blue zones in (right), corresponding to coverage by buildings, demonstrates that our approach can cope also with this kind of occlusion.

\subsection{Stability Through Different Environments}
\begin{SCfigure}
\centering
\includegraphics[width=0.45\linewidth,trim={25mm 0mm 30mm 10mm}, clip]{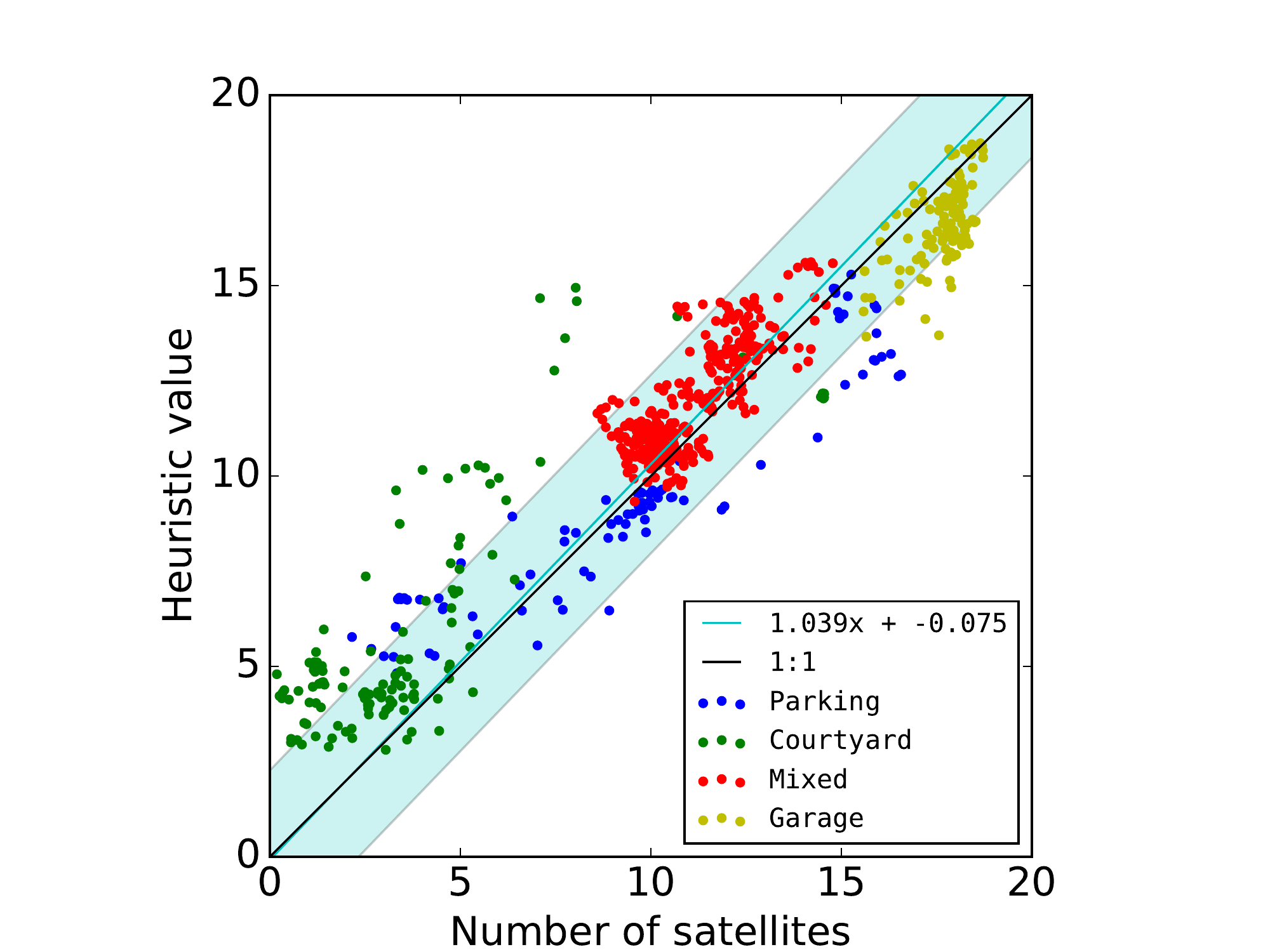}
\caption{Relation between satellite visibility and the number of satellites seen.
Each of the trajectories has been weighted by the number of points and from that we fitted a polynomial equation.
This equation (cyan line) is compared with a 1:1 black line.
The cyan area is the mean squared error from the fitting.
Since the number of satellites has an influence on the \ac{DOP} \cite{DOP}, the model can help determine the confidence in the positioning output.
}
\label{fig:violin}
\end{SCfigure}

To further illustrate the ability of our model in \autoref{eq:modelTransition} to cope with a wide variety of environments, we show the relationship between the actual and predicted number $\hat{v}$ of visible satellites. 
The predictions are for the four trajectories \texttt{Garage-1}, \texttt{Mixed}, \texttt{Parking} and \texttt{Courtyard} that were used to identify the parameters $\alpha$, $\beta$ and $\gamma$.
The plot in \autoref{fig:violin} shows that our model is indeed able to predict the number of visible satellites with a reasonable error, for all four trajectories. Moreover, it is able to cope with a large range of satellite visibility (0-20).

\subsection{Validation of the Predicted Coverage}

\begin{SCfigure}
\centering
\includegraphics[height=100px, trim={15mm 5mm 5mm 5mm}, clip]{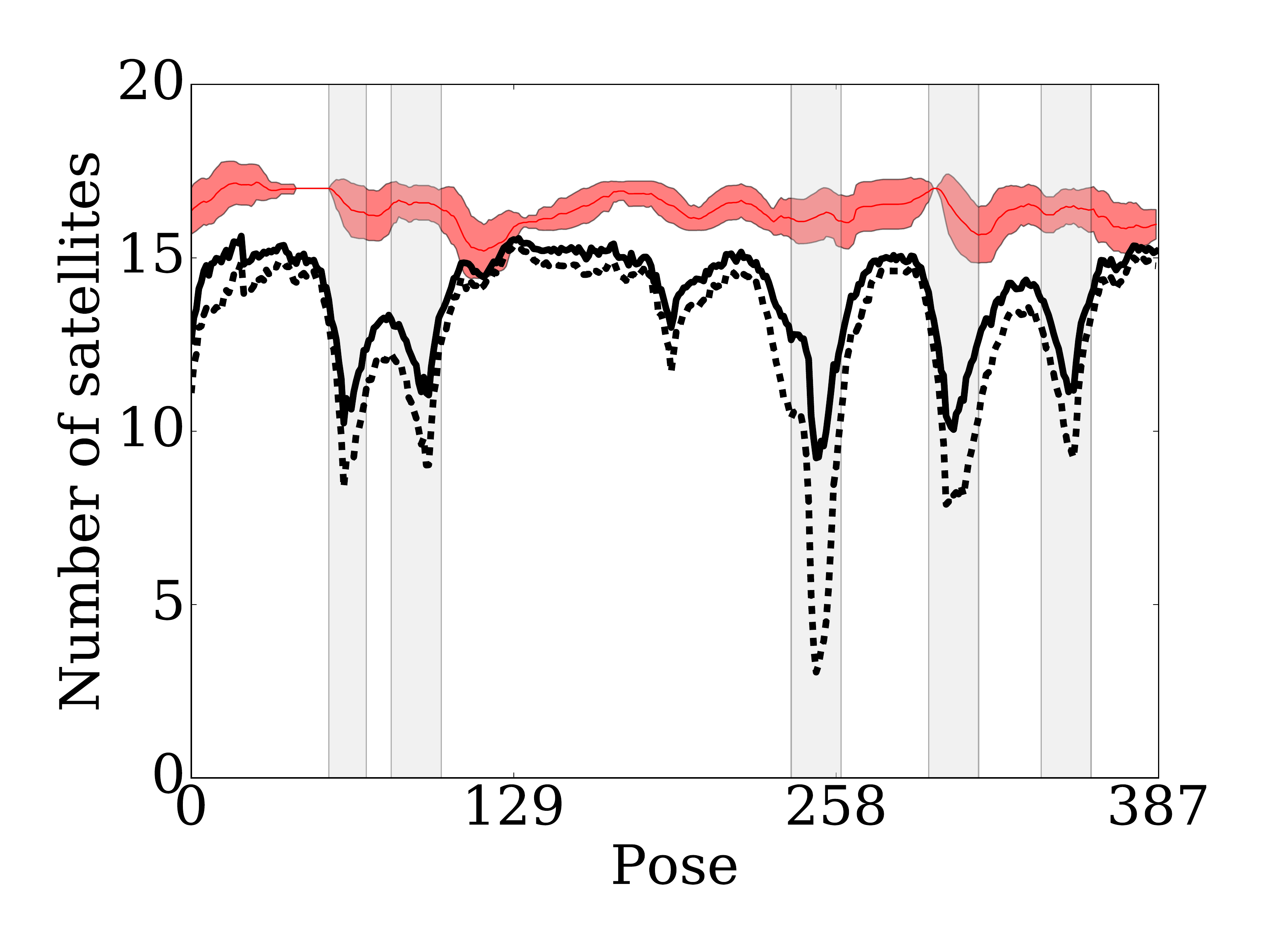}
\includegraphics[height=100px, trim={5mm 5mm 5mm 5mm}, clip]{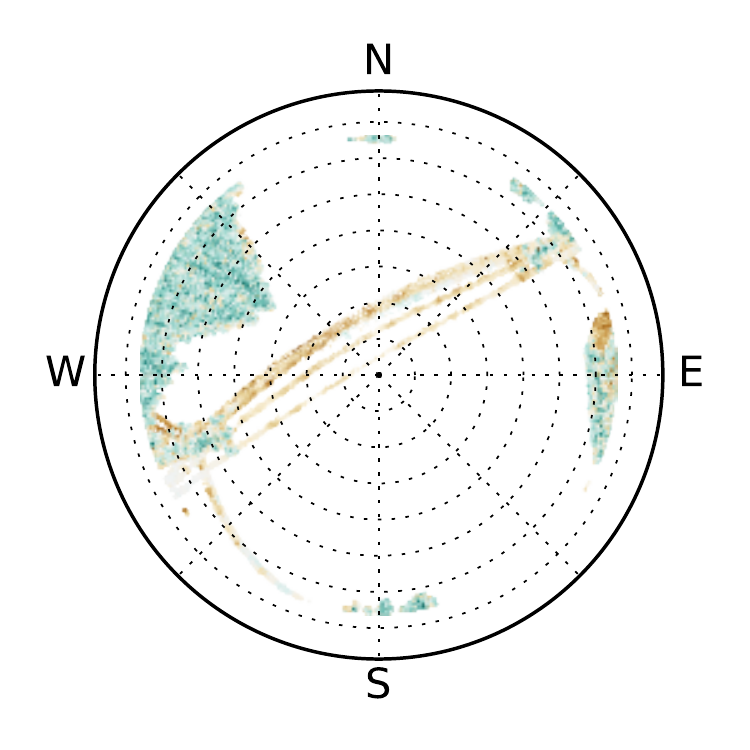}
\caption{
\emph{Left}: Validation trajectory performance with the areas with power-lines in gray.
\emph{Right}: Point cloud with view of the power-lines that are identified as structured points.
}
\label{fig:garage2_perf}
\end{SCfigure}

Since \texttt{Garage-2} is the same environment, the performance of our model should be similar.
The \autoref{fig:garage2_perf} shows the predicted number $\Hat{v}$ of visible satellite, along the trajectory.
Just like the previous ones, our model tends to underestimate the number of visible satellites.
Again, the model of \citet{GPS_remove_3d_points} performs worse than ours, with a much more significant underestimation.

\begin{SCfigure}
\centering
\includegraphics[height=90px]{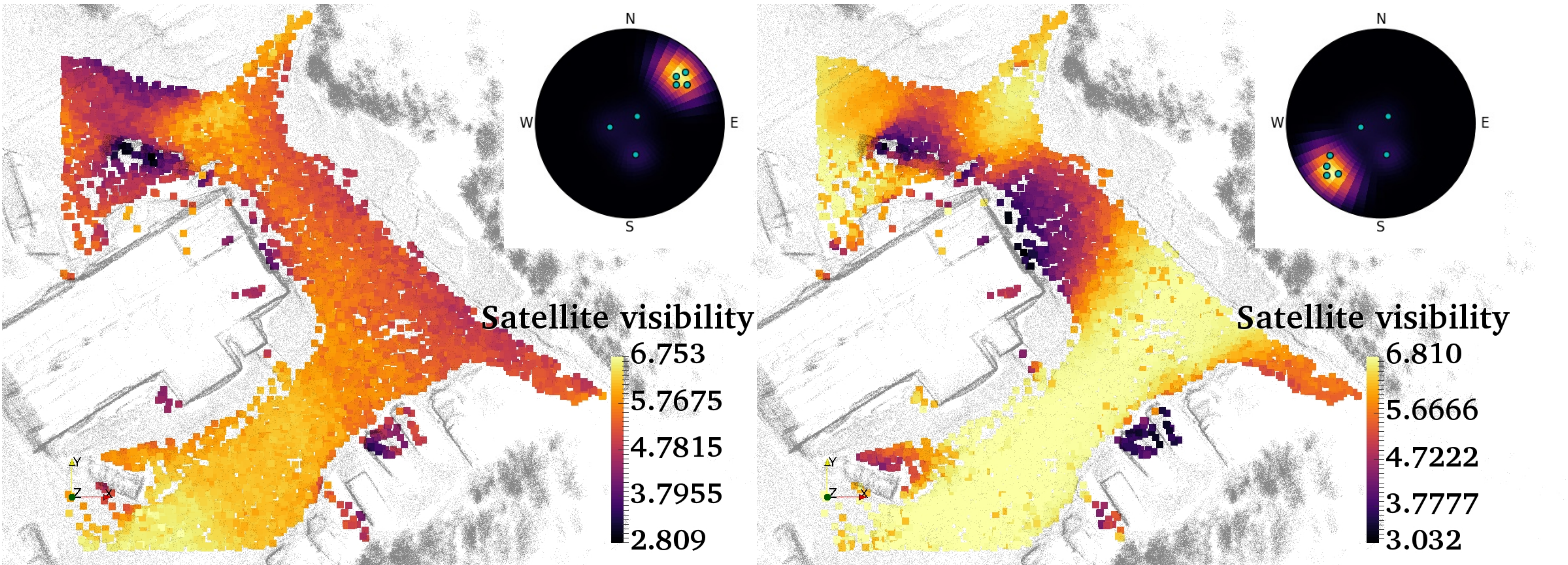}

\caption{
Predicted coverage on the ground plane.
\emph{Left}: $\mathbf{M_v}$ with satellite signal from around the north east.
\emph{Right}: $\mathbf{M_v}$ with satellite signal from around the south west.
}
\label{fig:sat_visibility}
\end{SCfigure}

Given our model, a query time and a 3D map, we can now predict ahead of time what would be the positions for a \ac{GNSS} receiver. Such a \emph{visibility map} $\mathbf{M_v}$ is depicted in \autoref{fig:sat_visibility}.
By changing the orientation of the satellites in the sky, we can see the effect it has on the map  $\mathbf{M_v}$.
We can see that in certain areas, such as near buildings or trees, the number of expected visible satellites is reduced. This visibility map $\mathbf{M_v}$ could be given to a path planning algorithm that minimizes the probability of getting lost \cite{PathPlanningNickRoy}.

\subsection{Histogram resolution on occupancy}
\begin{SCfigure}
\begin{minipage}[t]{0.6\textwidth}
\includegraphics[width=0.32\linewidth, trim={4mm 5mm 5mm 4mm}, clip]{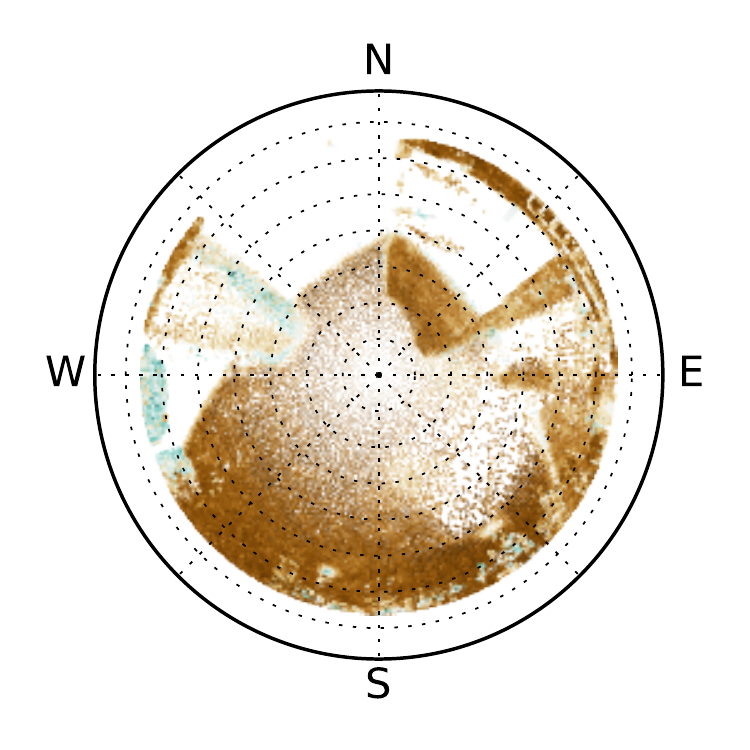}
\includegraphics[width=0.32\linewidth, trim={4mm 5mm 5mm 4mm}, clip]{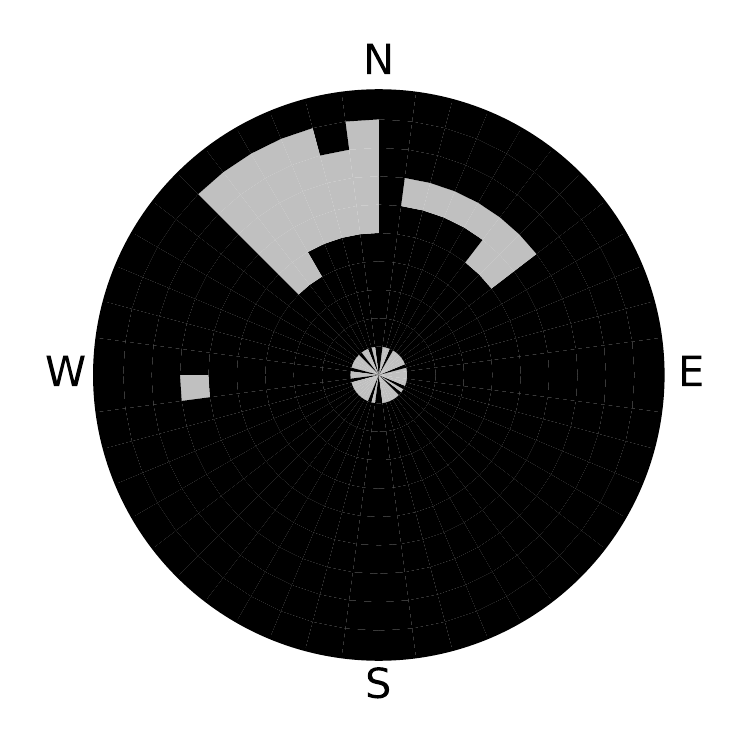}
\includegraphics[width=0.32\linewidth, trim={4mm 5mm 5mm 4mm}, clip]{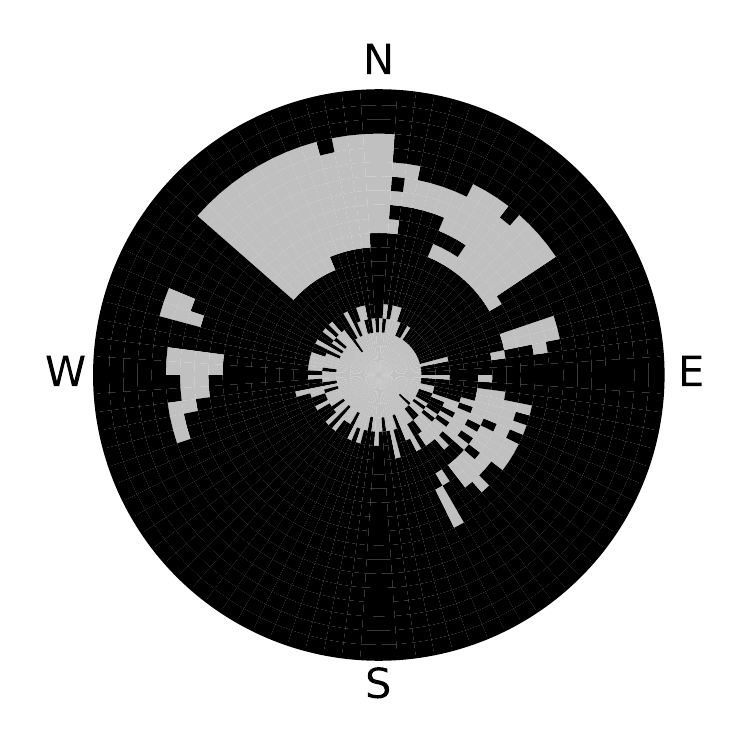}
\includegraphics[width=0.32\linewidth, trim={11mm 5mm 12mm 0mm}, clip]{figs/fig4/colbar/points_colobar.pdf}
\includegraphics[width=0.32\linewidth, trim={11mm 5mm 12mm 0mm}, clip]{figs/fig4/colbar/occupancy_bar.pdf}
\includegraphics[width=0.32\linewidth, trim={11mm 5mm 12mm 0mm}, clip]{figs/fig4/colbar/occupancy_bar.pdf}

\caption{Effect of the resolution on the binary mask $\bm{B}$.
\emph{Left}: point cloud of a roof and a pillar.
\emph{Middle}: Occupancy histogram with a resolution of \SI{7.5}{\degree} $\times$ \SI{9}{\degree}.
\emph{Right}: Occupancy histogram with a resolution of \SI{3.75}{\degree} $\times$ \SI{4.5}{\degree}.}
\label{fig:histgram_resolution}

\end{minipage}
\end{SCfigure}

One encountered difficulty is that a planar surface may not produce as many points as a tree would.
Because of this, the histogram bins contain fewer values to work with and so the precision decreases as shown in \autoref{fig:histgram_resolution}. 
If the resolution is too high, some parts of the map may be classified as unoccupied in the binary mask $\bm{B}$ (e.g., a wall of a building which is not dense enough).
The signal is then incorrectly let through by the model.

A possible solution to this problem is reducing the size $d_\text{box}$ of the bounding box of the octree filter or using ray tracing handling octree directly.
This adjustment would increase the amount of points avoiding the original cause of the problem.

\section{Conclusion}
\label{sec:Conclusion}
In this paper, we proposed a new model for estimating the number of visible \ac{GNSS} satellites, using a 3D point cloud map and future satellite constellation information. Our model takes into account both absorbing (e.g. trees) and occluding (e.g. building) structures, by only relying on the distribution and density of the 3D map points. 
We also designed a flexible data gathering platform that is able to function in harsh Canadian winters.
To estimate and validate parameters of the model, we moved our platform along several trajectories in diverse environments that featured dense forests and high buildings.
Contrary to a pure masking approach based on the line-of-sight, our model was able to successfully predict the number of visible satellites in all these environments. 
The prediction can thus be useful for trajectory planning algorithms which rely on localization precision---the number of viable satellites influences the \ac{DOP}.
Future work includes explicit \ac{DOP} prediction based on the geometry of the remaining satellites and their individual predicted \ac{SNR}.

\begin{acknowledgement}
We thank everyone at the Montmorency forest for their help,
and their equipment.
This research was supported by the Natural Sciences and Engineering Research Council of Canada (NSERC) through grant CRDPJ 511843-17 BRITE (bus rapid transit system) and FORAC.
Finally, we also thank Maxime Vaidis and Charles Villeneuve for their help with the snowmobile.
\end{acknowledgement}

\bibliography{references}

\end{document}

%% file: preamble.tex
\usepackage[l2tabu,orthodox]{nag}

\usepackage[sort, square,numbers]{natbib}
\usepackage[ pdftex,colorlinks]{hyperref} %

\usepackage{layouts} %

\usepackage[printonlyused]{acronym}
\usepackage{siunitx}
\usepackage[all]{nowidow}

\usepackage[dvipsnames, table]{xcolor}

\usepackage{lipsum}

\usepackage{verbatim}

\usepackage{mathtools}

\usepackage{subcaption}

\usepackage{sidecap}

\usepackage{pdflscape}

\usepackage{pifont}

\usepackage[inline]{enumitem}

\usepackage[pdftex]{graphicx}
\usepackage{epstopdf}

\usepackage{import}

\usepackage{float}

\usepackage[font=footnotesize]{caption}

\graphicspath{{./latexGoodPractices/}}

\usepackage{booktabs}

\usepackage{tabu}

\usepackage{supertabular}

\usepackage{rotating,tabularx}

\usepackage{makecell}

\usepackage{amssymb,amsfonts,amsmath,amscd}

\usepackage{bm}

\newcommand{\bbm}{\begin{bmatrix}}
\newcommand{\ebm}{\end{bmatrix}}